\documentclass[10pt, a4paper]{article}
\usepackage{lrec}
\usepackage{multibib}
\newcites{languageresource}{Language Resources}
\usepackage{graphicx}
\usepackage{tabularx}
\usepackage{soul}

\usepackage{pdfpages}
\usepackage{epstopdf}
\usepackage{newunicodechar}
\usepackage[X2, T1]{fontenc}
\usepackage[utf8]{inputenc}
\usepackage{soul}
\usepackage{longtable}
\usepackage{multirow}
\usepackage{xstring}
\usepackage{tipa}
\usepackage{subfig}
\usepackage[colorlinks]{hyperref}
\usepackage{indentfirst}
\usepackage{placeins}
\usepackage{dblfloatfix}

\newcommand{\nlang}{67 }
\newcommand{\nfam}{28 }

\newunicodechar{Ә}{{\fontencoding{X2}\selectfont\CYRSCHWA}}

\usepackage[font=small]{caption}

\usepackage{color}
\definecolor{darkblue}{rgb}{0.0, 0.0, 0.55}
\hypersetup{
    colorlinks = True,
    urlcolor = darkblue,
    linkcolor= darkblue,
    citecolor=.}

\title{Building and curating conversational corpora for
diversity-aware language science and technology}

\name{Andreas Liesenfeld, Mark Dingemanse \textsuperscript{*}\thanks{\hspace{-18pt} * Both authors contributed equally.}}

\address{Centre for Language Studies \\
         Radboud University, Netherlands \\
         andreas.liesenfeld@ru.nl, mark.dingemanse@ru.nl}

\abstract{
We present an analysis pipeline and best practice guidelines for building and curating corpora of everyday conversation in diverse languages. Surveying language documentation corpora and other resources that cover \nlang languages and varieties from \nfam phyla, we describe the compilation and curation process, specify minimal properties of a unified format for interactional data, and develop methods for quality control that take into account turn-taking and timing. Two case studies show the broad utility of conversational data for (i) charting human interactional infrastructure and (ii) tracing challenges and opportunities for current ASR solutions. Linguistically diverse conversational corpora can provide new insights for the language sciences and stronger empirical foundations for language technology. 
\newline  }

\begin{document}

\maketitleabstract

\begin{figure}[!b]
\begin{center}
\includegraphics[width=\textwidth]{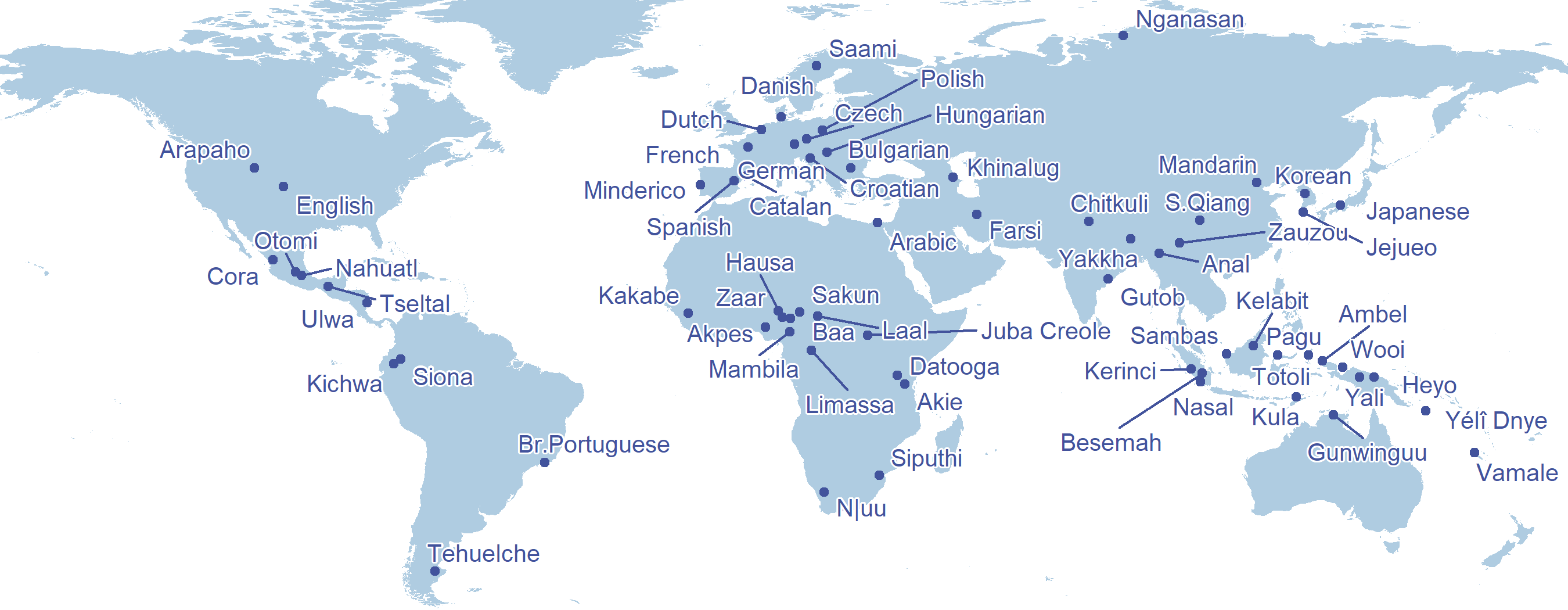}
\onecolumn
\caption{Languages currently featured in the dataset plotted by geographic location of (one of their) speech communities. Coordinates from Glottolog \protect\cite{hammarstromGlottologGlottologGlottolog2021}; full list in Appendix.}
\twocolumn
\label{fig.1}
\end{center}
\end{figure}

\section{Introduction}
\noindent
Language resources that capture language use in its natural habitat of social interaction are rare despite the obvious merits of studying the very environment where we all learn language and use it everyday  \cite{schegloff_interaction_2006}. There are multiple reasons for this. Linguists have been trained to look the other way when it comes to what is considered mere performance \cite{boeckxLanguageCognitionUncovering2010}. Collecting this kind of data requires one to venture out of lab settings and other controlled environments \cite{enfieldDoingFieldworkBody2013}, and transcribing it is resource-intensive \cite{himmelmannMeetingTranscriptionChallenge2018}. These obstacles are compounded by the fact that most NLP work focuses on a handful of well-studied languages \cite{joshiStateFateLinguistic2020,blasiSystematicInequalitiesLanguage2021}. However, under the auspices of various language documentation projects, language resources have been collected in more and more communities across the world \cite{seifartLanguageDocumentationTwentyfive2018}, and these often include at least some conversational data. We argue such corpora harbour important insights for language science and technology.

\begin{figure}[hbt!]
\begin{center}
\includegraphics[scale=0.75]{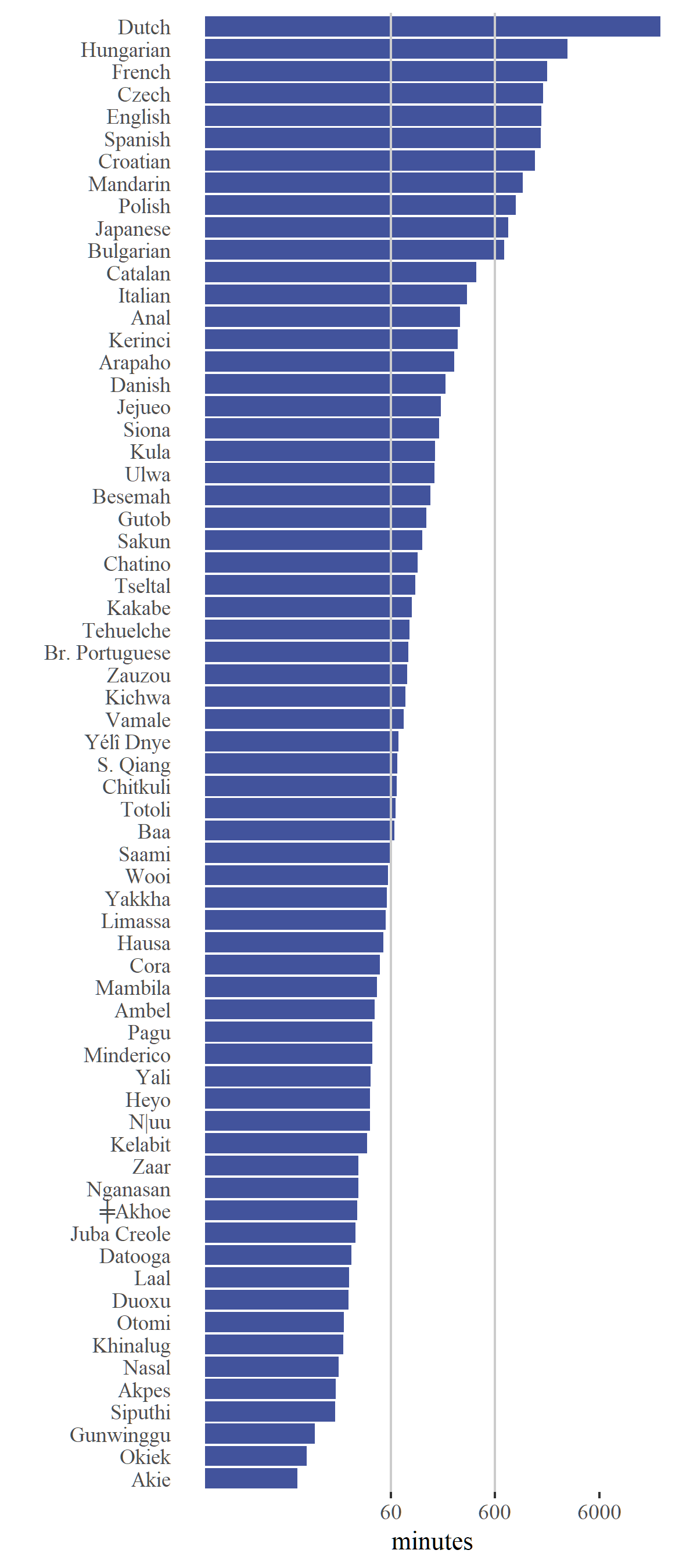} 
\caption{Language and corpus size in minutes (log scale).}
\label{fig.2}
\end{center}
\end{figure}

In this paper we describe efforts to build an open and reproducible pipeline for collating and curating corpora of conversational speech. We demonstrate use of the pipeline for a growing collection of corpora covering at least \nlang languages of \nfam phyla (Figure \ref{fig.1}). Around 75\% of the corpora are sourced from existing language documentation projects. The remaining 25\% come from other language resource platforms made available to the research community. Investigating a larger slice of the world's linguistic diversity can strengthen the empirical foundation of the language sciences and foster diversity-aware language technologies. 
\newpage
We publish a \href{https://osf.io/cwvbe/}{repository} that collects information on content and availability of corpora of conversational interaction across many languages. Currently, the curated collections amount to over 70 open datasets, representing over 700 hours of social interaction, 1.3 million annotations and over 8 million words (Table \ref{tab.1}, Figure \ref{fig.2}). We also publish Python and R scripts to assess the content and quality of conversational corpora. In this paper, we detail the data analysis pipeline and formulate best practices for corpus creators to optimally prepare corpora for interoperability. We also indicate some directions for research using such corpora in the form of two case studies, one aimed at linguistic typology and the other at speech technology. 

\begin{table}[!h]
\begin{center}
\begin{tabularx}{\columnwidth}{|l|X|}

      \hline
      Languages & \nlang\\
      \hline
      Phyla & \nfam\\
      \hline
      Hours of recordings & 705\\
      \hline
      Annotations (turns) & 1.3 million\\
      \hline
     Tokens (estimate) & 8 million\\
      \hline
\end{tabularx}
\caption{Dataset size overview.}
\label{tab.1}
 \end{center}
\end{table}

For sourcing conversational corpora, we have used the following three criteria. First, the resource should be \textit{maximally naturalistic}, capturing informal, unscripted interaction between two or more participants. We assess this by looking at the dynamics of turn-taking and timing, aiming to select corpora or sub-corpora characterized by free-flowing, unscripted interaction. Second, we aim for a \textit{maximally diverse} dataset that covers many languages and phyla beyond the usual handful of languages (mostly Indo-European) that still make up the bunk of available datasets of conversational speech. Third, in order to foster open science and reproducible research, we privilege \textit{open} resources made available with informed consent and accessible free-of-charge to the research community. As such, the bulk of the data comes from language resource repositories, often related to language documentation projects: \href{http://www.elda.org}{ELDA Shared LRs}, \href{https://dobes.mpi.nl}{Dobes (The Language Archive)}, \href{https://www.elararchive.or}{ELAR}, \href{https://talkbank.org}{Talkbank}, \href{https://openslr.org}{OpenSLR}, and so on. In some cases, commercial platforms also host accessible data, e.g. \href{https://www.ldc.upenn.edu}{Linguistic Data Consortium (LDC)}. Other sources of conversational corpora are national research projects like \href{https://taalmaterialen.ivdnt.org/download/tstc-corpus-gesproken-nederlands/}{Spoken Dutch Corpus (CGN)}, \href{http://cass.lancs.ac.uk/cass-projects/spoken-bnc2014/}{Spoken British National Corpus}, \href{https://ccd.ninjal.ac.jp/csj/}{NINJAL in Japan} and 
\href{http://agd.ids-mannheim.de/folk.shtml}{FOLK in Germany}. However, some of these resources (e.g., Spoken BNC, NINJAL) do not provide turn-based timing information and were therefore not included. A complete and up to date list of all data sources can be found through \href{https://osf.io/cwvbe}{osf.io/cwvbe}.

\section{Parsing conversational corpora}
\noindent
Conversational corpora come in various representation formats and levels of transcription granularity. There is no one unified representation of talk that would equally satisfy the needs of researchers working in different corners of the language sciences, be it grammar writing, conversation analysis, or phonetics \cite{Ochs1979,boldenTranscribingResearchManual2015,couper-kuhlenInteractionalLinguisticsIntroduction2017}. As a result, textual representations of conversation come in a range of formats and with various layers of information. Many existing pipelines for working with corpora are built for textual data instead of time-aligned transcriptions of social interaction. Notable recent exceptions are the R packages \texttt{chattr} \cite{casillasAnalyzingContingentInteractions2021} and \texttt{act} \cite{ehmerActAlignedCorpus2021}, and a workflow for dealing with XML \cite{ruhlemannVisualLinguisticsPractical2020}. Less directly aimed at co-present interaction are ConvoKit \cite{changConvoKitToolkitAnalysis2020} and the DoReCo pipeline \cite{paschenBuildingTimeAlignedCrossLinguistic2020}, built primarily for research into word-level time-alignment. 

Despite the diversity in formats, some structural features are important for any corpus of conversation. These relate to the primary annotation level, the importance of timing and participation, the representation of supposedly marginal features, and the linking of annotations and source files. To discuss these in turn:

\begin{figure}[b!]
\begin{center}
\frame{\includegraphics[width=\columnwidth]{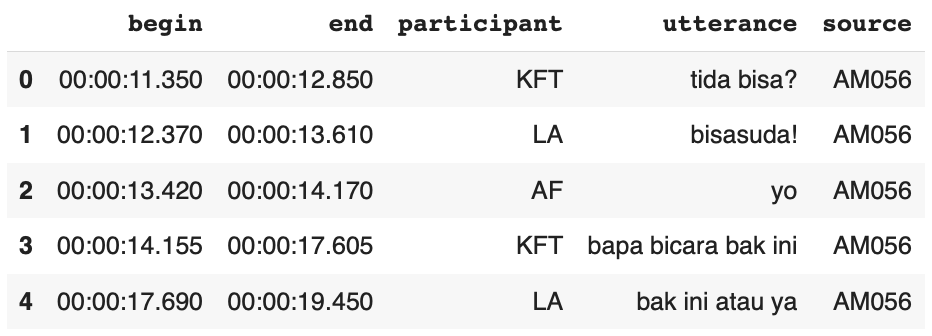}}
\caption{Example of the minimal viable format for conversational data (in dataframe format).}
\label{fig.3}
\end{center}
\end{figure}

\begin{itemize}

\item \textbf{The fundamental unit of organization: turns.} People organize interactions through turns at talk, which can be characterized as communicative moves that are recognizably complete for participants in interaction \cite{fordInteractionalUnitsConversation1996}. The idealised notion of `sentence' bears a complicated relation to empirically attested turns at talk  \cite{sacksSimplestSystematicsOrganization1974,kempsonLanguageMechanismsInteraction2016}. Somewhat closer to the level of turn in conversation is the notion of inter-pausal unit (IPU), which has the benefit of being automatable \cite{bigiSPPASMultiLingualApproaches2015}, though no automated method will capture the flexibility and fluidity of human judgments about turns and the social actions they implement. 

\item \textbf{Represent timing and participation.} People in interaction treat the timing and duration of utterances as orderly and meaningful \cite{sacksSimplestSystematicsOrganization1974}. They minimize gaps and overlaps \cite{stiversUniversalsCulturalVariation2009}, and are demonstrably sensitive to timing differences on the order of a few hundred milliseconds \cite{robertsIdentifyingTemporalThreshold2013}. The accurate representation of who said what and when exactly (with at least decisecond precision) is crucial to any work on human interaction. 

\item \textbf{Retain relevant details.} No element of talk can be treated as discardable a priori. Conversational transcripts contain complex turns but also one-word elements such as ``oh" or ``um" \cite{buschmeierCommunicativeListenerFeedback2018,williamsParsingParticlesWa2020}, and may also capture non-verbal conduct like as breaths, laughs, sighs, or coughs \cite{wlodarczakBreathingConversation2020,keevallikSoundsMarginsLanguage2020}. If the goal is to characterize, understand, and model turns at talk, then such elements should be represented where possible and relevant.

\item \textbf{Keep transcriptions and source files linked.} Since annotations and transcriptions are necessarily selective and made for a particular purpose, it is important to keep source data (audio, video, and any other streams of information like kinematics or eyegaze) closely linked to textual representations \cite{zimmermanAcknowledgmentTokensSpeakership1993}. This enables repeated inspection, opens up annotations and analyses to empirical scrutiny, and makes it possible to investigate aspects not captured by annotations.
\end{itemize}

Taking these properties into account, we define a minimal viable unified representation format for conversational speech in the form of a flat dataframe that features one participant turn per row and that has (at least) five columns: \texttt{begin} and \texttt{end} of the utterance, the \texttt{participant} producing the utterance, and the \texttt{utterance} content, and finally a \texttt{source} column that links the dataframe row to any corresponding media files (Figure \ref{fig.3}). If a source corpus contains additional information such as multiple scripts, translations, or annotation layers that capture lexical, phonetic, morphological or part-of-speech information, this is stored in additional columns. This way we ensure interoperability of core columns, but also keep additional information at hand on a per-corpus basis.

\begin{figure}[t!]
\begin{center}
\includegraphics[scale=0.35]{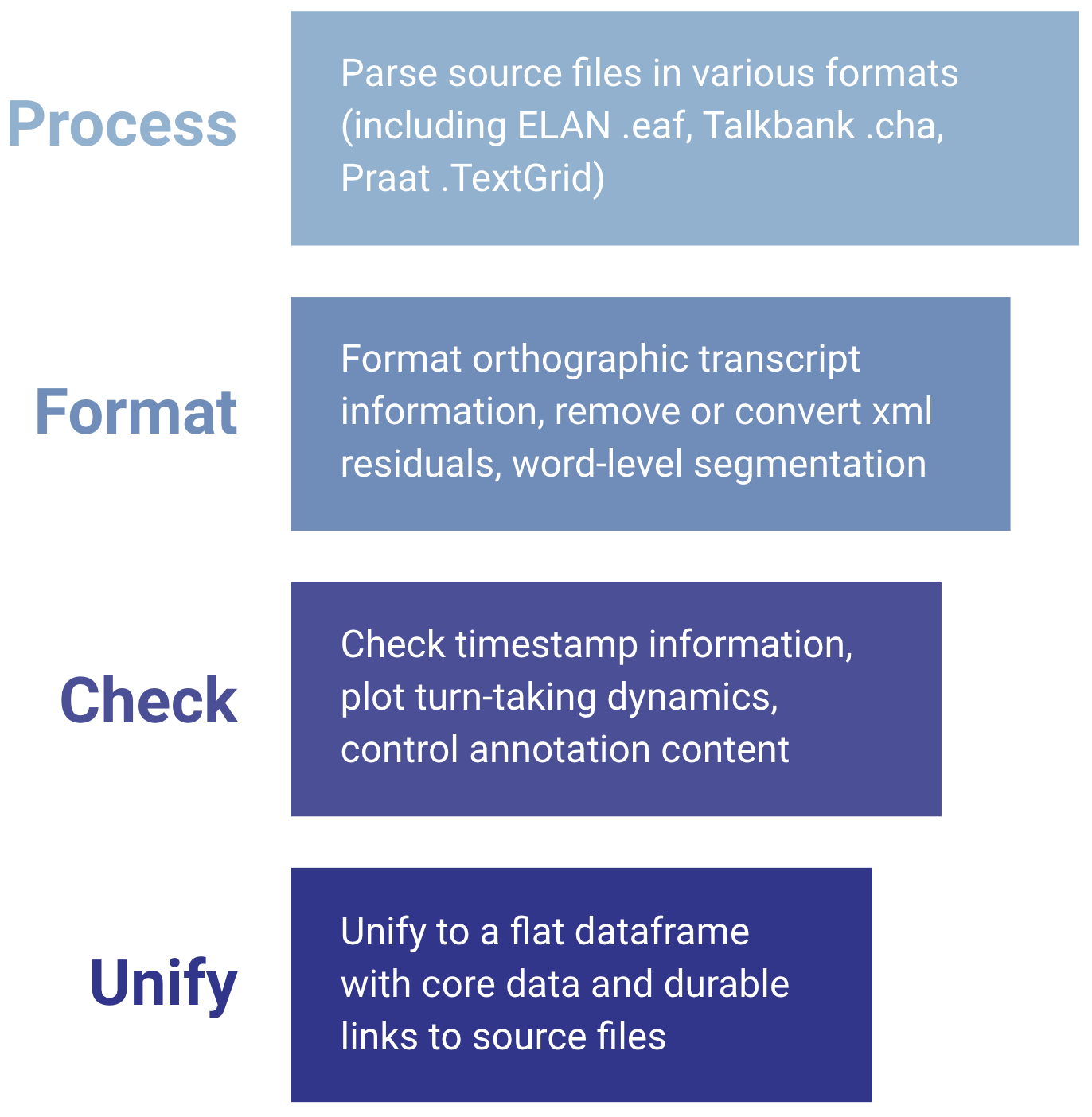} 
\caption{Overview of a four-step processing pipeline from raw transcription data to a minimal viable unified format.}
\label{fig.4}

\end{center}
\end{figure}

The minimal viable corpus format aims to make diverse conversational corpora amenable to fundamental and applied research by enabling a basic form of sequential analysis of talk as it unfolds over time. Given the critical importance of accurate timing data and high quality transcriptions in this process, our data analysis pipeline includes a number of quality control steps that focus on assessing the accuracy and quality of available conversational corpora. The pipeline can be broken down into 4 steps (Figure \ref{fig.4}). First we process the source transcription files by writing format-specific parsers. Then we format annotation metadata and content by converting timestamps to ms, clearing xml residuals and performing word-level segmentation. The extent of this task differs across languages: often splitting by whitespace is sufficient, but for other languages such as Chinese, we employ a parser. In a next step we check timing information quality and annotation content, using a common \texttt{[unk]} tag for missing annotations (see section 3 for details). Finally we unify the data by storing it in a flat dataframe with durable links to source files.

\begin{figure*}[!h]
\begin{center}
\includegraphics[width=\textwidth]{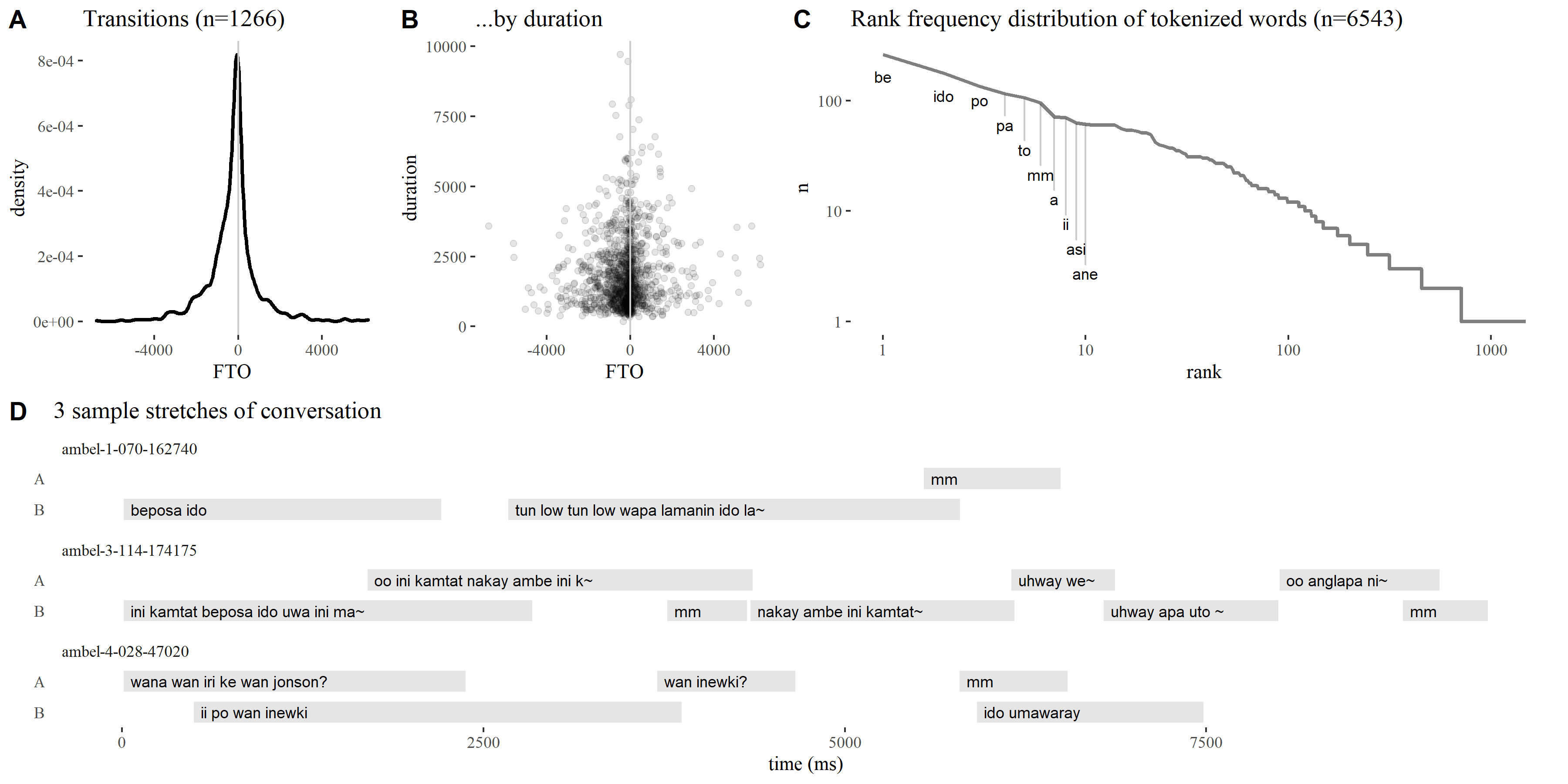}
\caption{Example of an assessment report for conversational data, here illustrated with data from Ambel \protect\citelanguageresource{arnoldDocumentationAmbelAustronesian2017}. \textbf{A}. Distribution of the timing of dyadic turn-transitions with positive values representing gaps between turns and negative values representing overlaps. This kind of normal distribution centered around 0 ms is typical; when corpora starkly diverge from this it usually indicates non-interactive data, or segmentation methods that do not represent the actual timing of utterances. \textbf{B}. Distribution of transition time by duration, allowing the spotting of outliers and artefacts of automation (e.g. many turns of similar durations). \textbf{C}. A frequency/rank plot allows a quick sanity check of expected power law distributions and a look at the most frequent tokens in the corpus. \textbf{D}. Three randomly selected 10 second stretches of dyadic conversation give an impression of the timing and content of annotations in the corpus. }
\label{fig.5}
\end{center}
\end{figure*}

\section{Curating conversational data}
\noindent
After parsing we perform several quality control steps to ensure that the corpus data is fit for inclusion in the curated collection. In particular, we verify annotation content, check source files, and assess timing information.

\textit{Annotation content.} We verify that annotation content is transcribed using a common orthographic format. For data processing reasons, we romanize all scripts (while preserving the original). Many corpora also contain annotations tagged for bodily conduct like laughter, breathing, or coughing. We aim to preserve as much of this information as possible, converting common tags to a unified tag format, and marking other such elements using square brackets.

\textit{Source files.} The baseline requirement of inclusion in the dataset is that audio data of reasonable quality exists that allows us to verify and unify transcription content using primary data. This is important to alleviate risks of grounding any subsequent analysis on transcriptions alone. We also record any other data streams (e.g. video, gaze), verify that every source reference in the data has a corresponding source file and manually inspect the overall recording quality. The resulting curated corpora can be used for automatic extraction of audio clips for further analysis, and for feature extraction of prosodic properties that are interactionally relevant like speech rate, intensity and pitch \cite{seltingProsodyActivitytypeDistinctive1996,wardBottomUpExplorationDimensions2012}.

\textit{Timing.} Given the importance of timing and participation, we aim to ensure that timing information is as complete and accurate as possible across the dataset. For the purpose of examining talk-in-interaction we define this as timestamps that accurately correspond to the beginnings and ends of conversational turns. This also includes the accurate identification of overlaps and gaps between turns. Quality control of these measures is done through a combination of manual inspection and quantitative measures, for instance by plotting the dynamics of turn-taking and timing (Figure \ref{fig.5}A-D). 

Doing this for over 50 corpora, we identified several recurring issues. One is incomplete transcriptions. We compute the relative annotation density in order to get a sense of the likelihood of missing annotations. Another issue is inaccurate timing information, which may arise from the use of automated transcription methods or shortcuts in manual annotation software. To diagnose such issues at scale, we generate an assessment report for every language (see Figure \ref{fig.5}) with information on turn transition timings, turn durations, rank-frequency distributions, and sample stretches of dyadic conversation. This enables a quick assessment of the relative precision and granularity of a corpus. For instance, timing inaccuracies are often recognizable by deviations from the expected normal distribution \cite{stiversUniversalsCulturalVariation2009}; and sample plots of conversations give an impression of interaction type and annotation content, including empty annotations. A codebase for generating such assessment reports is available in the \href{https://osf.io/cwvbe/}{repository}.

\section{Building corpora: best practices}
\label{recommendations}
\noindent
There is still a relative dearth of conversational corpora, and most of the world's linguistic diversity remains underrepresented. The reasons for this are varied and include the fact that conversational data is seen as hard to collect and even harder to analyse, with the language sciences preferring to focus on sanitized versions of linguistic structure and behaviour and language technology similarly focusing on text and speech data that is pristine and can be easily processed. 

Broader representation is important for communities and heritage users of languages, who have long been underserved by monologic textual materials that make it hard to get a taste of what it is like to use a given language in face-to-face interaction \cite{ameryPhoenixRelicDocumentation2009}. But it is also important for scientific purposes, as every language offers its own contribution to the tapestry of unity and diversity that characterizes our common cultural heritage \cite{birdDecolonisingSpeechLanguage2020}.

Based on our experience building and processing conversational corpora, here we provide some recommendations for \textit{minimally useful conversational corpora}. Good corpora can often serve multiple purposes, from language learning to linguistic research \cite{enfieldDoingFieldworkBody2013} and from comparative investigations to supplying training data for NLP purposes. Such corpora have the following properties:

\begin{itemize}
\item It is collected and archived with consent of participating language users and the relevant community leaders
\item It contains audio and/or video recordings of everyday face to face interaction among multiple participants
\item It is time-aligned at the level of conversational turns, with turns allotted to participants
\item It is transcribed in a way that provides access to the linguistic material beyond the audio/video
\end{itemize}

There are good guides for building and transcribing corpora \cite{allwoodMultimodalCorpora2008,paschenBuildingTimeAlignedCrossLinguistic2020}, and increasingly, tools are available to transcribe audio and video data \cite{birdSparseTranscription2021,dingemanseHighSpeedTranscription2012} and enrich annotations \cite{umairGailBotAutomaticTranscription2021,zahrerBuildingAutomaticTranscription2020}.

\begin{figure}[!t]
\textbf{\textsf{A}}

\includegraphics[width=\columnwidth]{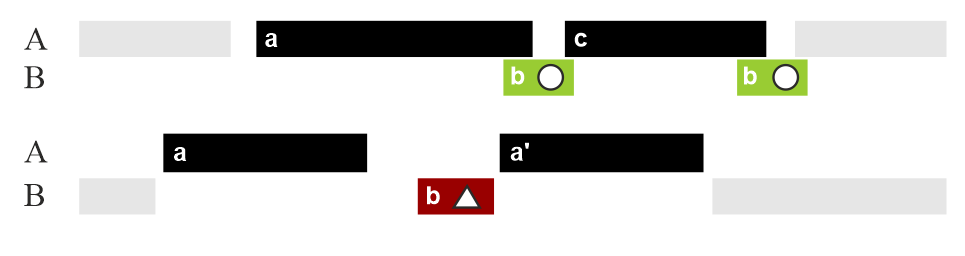}

\textbf{\textsf{B}}

\includegraphics[width=\columnwidth]{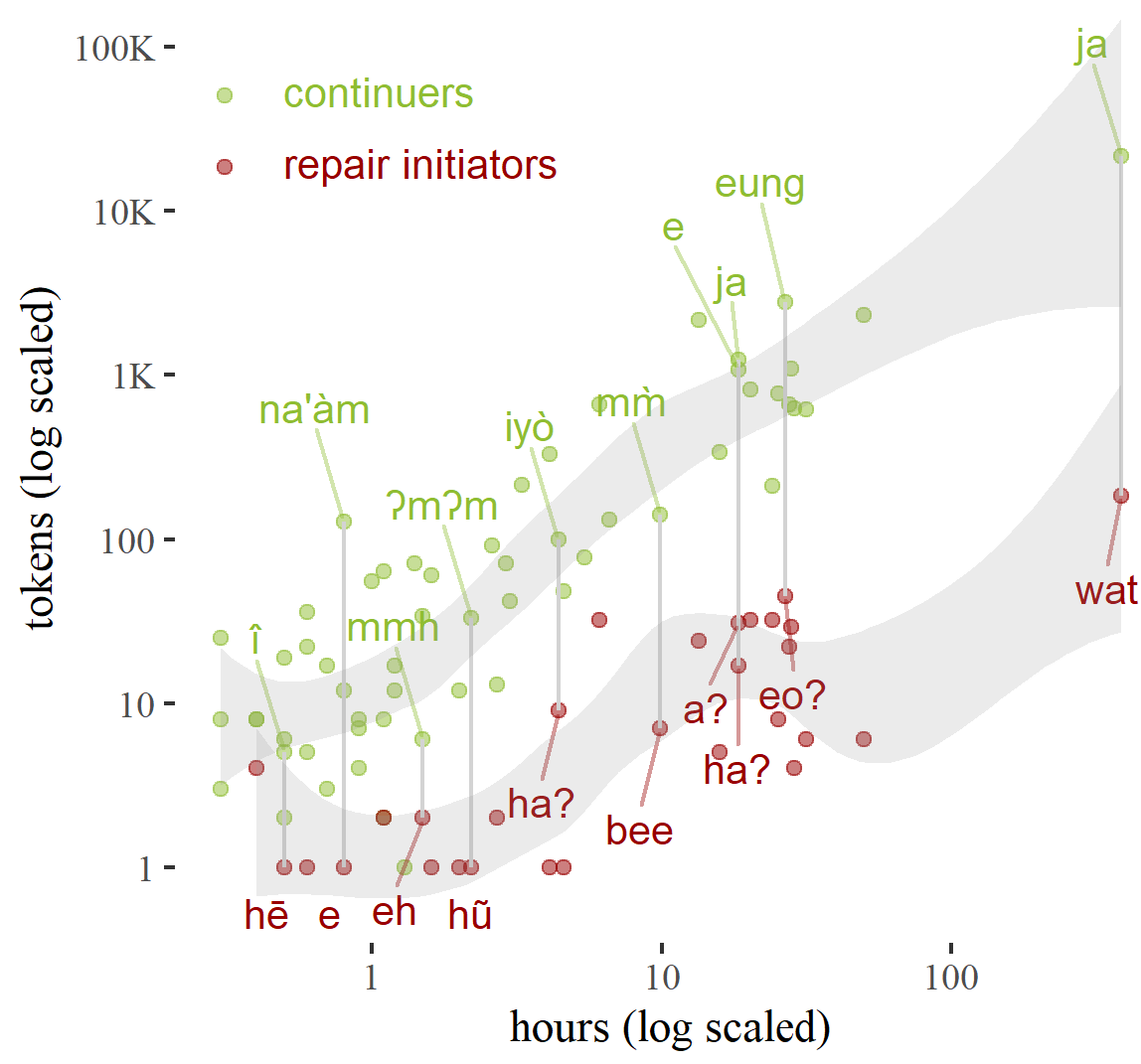}

\caption{\textbf{A}: Typical sequential structures for continuers versus repair initiators. Continuers are recurring items found in alternation with unique turns (\textit{a, c}). Repair initiators are recurring items found between a unique turn \textit{a} and its near-copy \textit{a'}. \textbf{B}: Prevalence of sequentially identified candidate continuers and repair initiators, demonstrating the potential of using sequential patterns to identify them in language-agnostic ways. Most frequent formats exemplified in 10 languages (9 phyla), from left to right: \textdoublebarpipe{}Akhoe Hai\textpipe \textpipe om, Hausa, Tehuelche, Gutob, Kerinci, Siwu, Mandarin, German, Korean, Dutch.}
\vspace{-30pt}
\begin{center}
\label{fig.6}
\end{center}
\end{figure}

\section{Exploring conversational corpora}
\noindent
To showcase the utility of conversational corpora, we provide two case studies. The first is focused on linguistic typology and investigates the use of conversational data for comparative studies of language structure. The second is focused on language technology and compares conversational corpora to typical speech recognition data. 

\begin{figure*}[!b]
\begin{center}

    \centering
    \subfloat{\includegraphics[width=5.2cm]{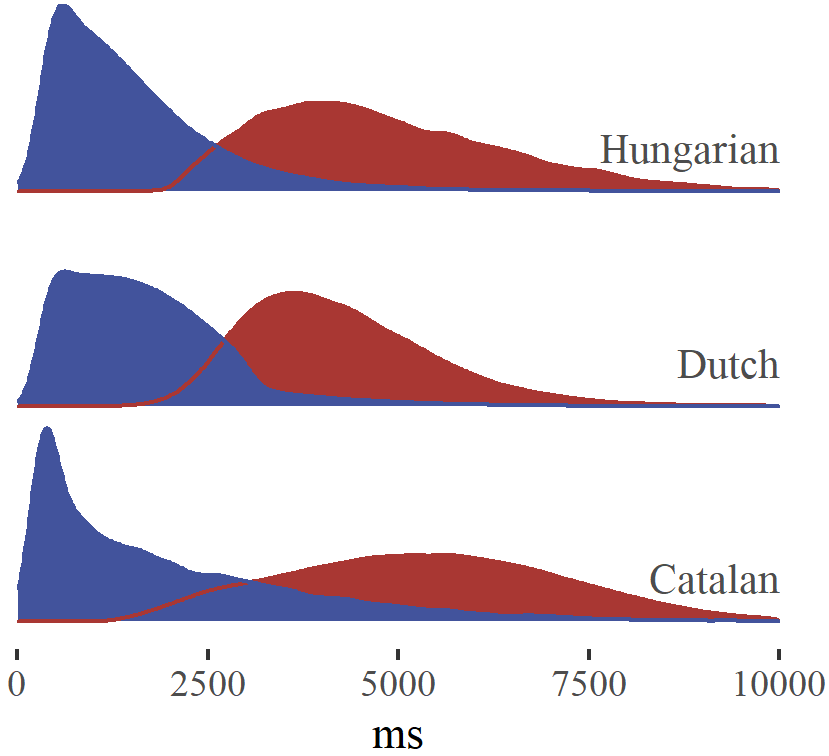}}
    \qquad
    \subfloat{\includegraphics[width=11.2cm]{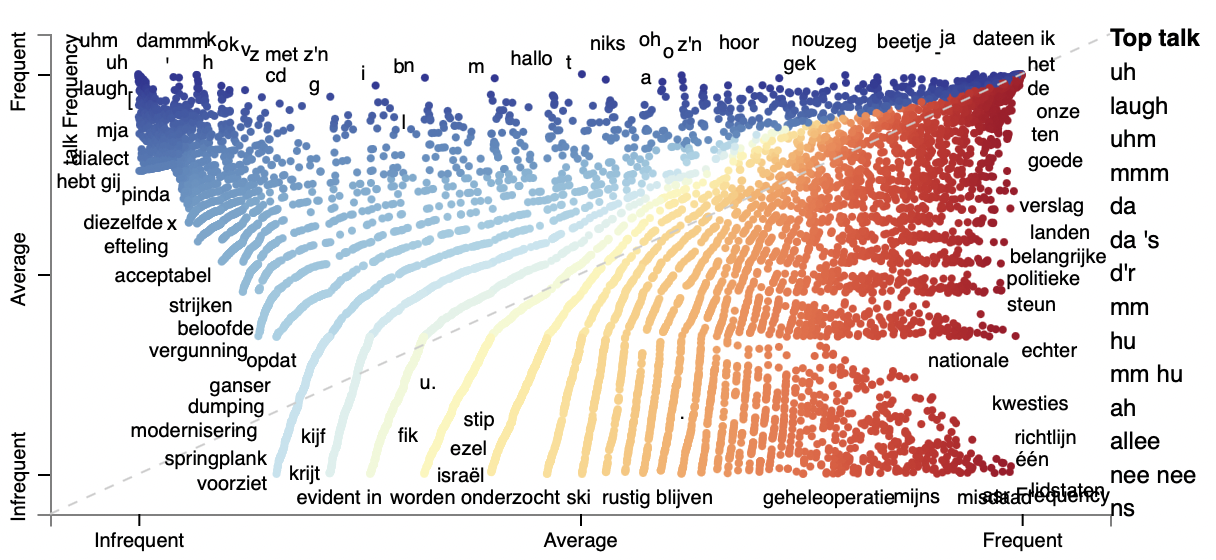}}

\caption{\textbf{L}: Distributions of durations of utterances and sentences (in ms) in corpora of informal conversation (blue) and CommonVoice ASR training sets (red) in Hungarian, Dutch, and Catalan. Modal duration and annotation content differ dramatically by data type: 496ms (6 words, 27 characters) for conversational turns and 4642ms (10 words, 58 characters) for ASR training items. \textbf{R}: Visualization of tokens that feature more prominently in conversational data (blue) and ASR training data (red) in Dutch. Source data: 80k random-sampled items from the Corpus of Spoken Dutch \protect\citelanguageresource{taalunieCorpusGesprokenNederlands2014} and the Common Voice corpus for automatic speech recognition in Dutch \protect\citelanguageresource{ardila2020common}, based on Scaled F score metric, plotted using \textit{scattertext}    \protect\cite{kesslerScattertextBrowserBasedTool2017}}
\label{fig.7}
\end{center}
\end{figure*}
\subsection{Continuers and repair initiators}
\noindent
One aspect of interactive language use that should be of considerable interest to language science and technology is the common occurrence of linguistic devices with primarily interactional functions \cite{allwoodSpeechManagementNonwritten1990,norrickInterjectionsPragmaticMarkers2009,liesenfeldCantoneseTurninitialMinimal2019}. Consider two stand-alone turn formats that are particularly frequent yet have very different functions (Figure \ref{fig.6}A). A \textit{continuer} signals an understanding that the other party is producing a series of turns and an expectation that more is coming; a \textit{repair initiator} signals a request for clarification and requires both parties to halt the conversation and interactively resolve the communicative trouble, often with a redoing of the prior turn as a result. 

If these items occur at all in text data, they are divorced from their interactional context; indeed, they are underrepresented even in scripted conversations \cite{prevot_grouping_2018,prevot_should_2019}, making them hard to identify a priori by systems that have access to form alone \cite{bender_climbing_2020}. How then can we identify them for cross-linguistic comparison and work towards their naturalistic implementation in cross-linguistically informed language technology?

The solution is to think in terms of the sequential structure of interaction \cite{jeffersonSequentialAspectsStorytelling1978,couper-kuhlenInteractionalLinguisticsIntroduction2017,dingemanseTextTalkHarnessing2022}. Using only sequential and frequency information, we can define the prototypical continuer as a recurrent turn format that occurs in alternation with near-unique turns by another speaker, and the prototypical repair initiation as a recurrent turn format found between a near-unique turn and its near-copy (Figure \ref{fig.6}A). Implementing these language-agnostic definitions as a sequential search template and using a normalised Levenshtein distance of $<$ 0.20 to identify near-similar turns, we are able to identify candidate continuers and repair initiators across corpora (Figure \ref{fig.6}B). Though identified in a fully language-agnostic way, the resources identified here appear to fit available linguistic descriptions quite well. For instance, in the repair initiators we recognize the `huh?'-like interjections and `what?'-based question words known from prior work in pragmatic typology \cite{enfieldHuhWhatFirst2013}, and among the continuers we find a similar mix of minimal `mhm'-like interjections and affirmative answers \cite{dingemanseInterjections2021}. 

The sequential search method is susceptible to fluctuations in corpus size and its interpretation for specific corpora will always require careful qualitative work. However, it can be used to quickly gauge the likely forms of key interactional tools, and can provide a lower bound on the amount of data needed to identify interactional tools in conversational corpora. For instance, in this case, we may conclude that an hour of conversational data can be sufficient to identify the most important interactional tools. More generally, corpora like this can also inform careful interactional linguistic work that respecifies traditional linguistic concepts \cite{ozerovThisResearchTopic2022}. Interactional data analysed in terms of sequential and distributional properties is likely to be a crucial element in the toolbox of conversational NLP, providing definitions and distinctions that are easily lost in tokenisation, part-of-speech tagging and machine translation. 

\subsection{Conversational vs. ASR corpora}
\noindent
While corpora of informal conversation are relatively rare, \textit{speech corpora} are standard fare in language technology and particularly automatic speech recognition (ASR). Indeed readers familiar with work in this domain may wonder why we have not included such speech resources, many of which are openly available. 

The most important reason for this is that few if any ASR corpora are based on conversational interaction. Instead, ASR corpora usually are derived from carefully read speech samples, whether audiobooks as in Librispeech \citelanguageresource{panayotovLibrispeechASRCorpus2015}, elicited text prompts from Wikipedia as in CommonVoice \citelanguageresource{ardila2020common}, or European Parliament proceedings as in VoxPopuli \cite{wangVoxPopuliLargeScaleMultilingual2021}. This makes ASR corpora very useful for recordings that share key features with the training data (i.e., audio that is monologic and comes in fairly long sentences). But the character of the training data may limit the quality and application potential of ASR in other domains such as conversational user interfaces.

\begin{figure}[hbt!]
\begin{center}
\includegraphics[scale=0.85]{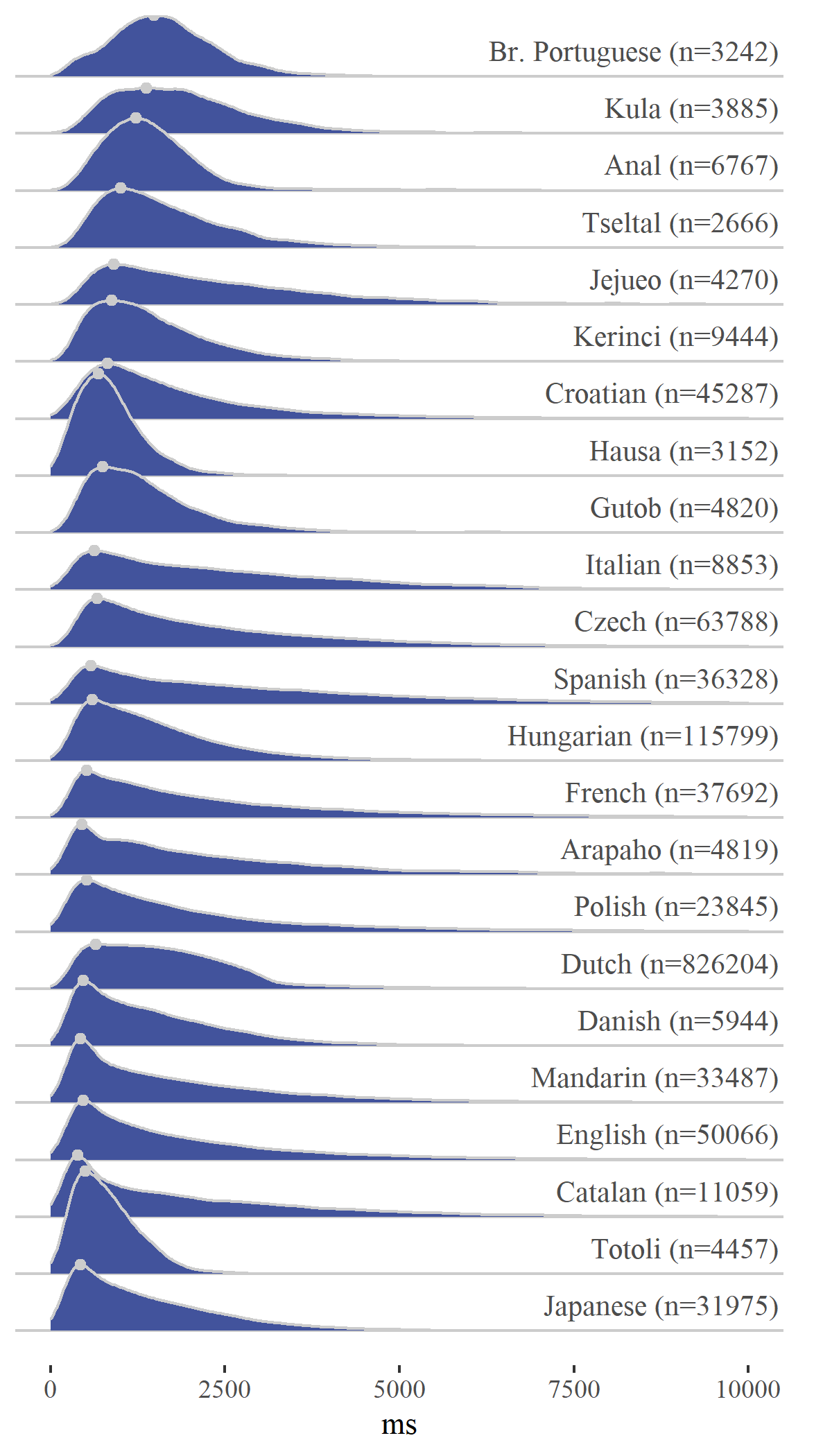} 
\caption{Distributions of turn durations (ms) in conversational corpora for 22 languages with at least 2500 turns. Across all languages (950k turns in total), the modal duration is 495ms (mean 1760ms, sd 1414ms). Recall that the modal duration of training items in most ASR corpora is an order of magnitude larger.}
\label{fig.8}
\end{center}
\end{figure}

Some of the differences are obvious (Figure \ref{fig.7}). First of all, ASR corpora come in chunks that are optimally sized for current ASR training solutions, meaning they are rarely longer than 35 seconds and rarely shorter than 2 seconds \citelanguageresource{panayotovLibrispeechASRCorpus2015}. These chunks are sentences or fragments of monologic textual corpora. Conversational corpora on the other hand come in turns that are optimally sized (by communicating participants) for the delivery of social actions \cite{enfieldAnatomyMeaningSpeech2009}. Typical ASR training chunks are much longer than turns at talk in conversation. The modal length of CommonVoice recordings for Hungarian, Dutch and Catalan is 4.6 seconds; for conversational corpora for these same three languages it is 0.5 seconds. The differences are striking enough that the overall distributions of utterances and sentences overlap only for a small part of the data (Figure \ref{fig.7}L). This pattern is not unique to the corpora compared here, or indeed to the CommonVoice dataset: the mean length of LibriSpeech English \citelanguageresource{panayotovLibrispeechASRCorpus2015} and of RuLS Russian sentences \cite{bakhturinaToolboxConstructionAnalysis2021} is also around 3 seconds. For comparison, in Figure \ref{fig.8} we provide the distributions of turn lengths in 22 conversational corpora from our dataset. This shows that conversational turns typically are at least twice as short as the typical sentences found in ASR training datasets.

Length may be the most obvious difference but it is not the main issue. The content of typical chunks differs quite a lot across corpus types. Figure \ref{fig.7}R shows how language differs according to corpus type for Dutch (comparing conversational data and the open CommonVoice dataset). The type of language most distinctive of the CommonVoice corpus mark its origin in parliamentary recordings, with formal words like ``verslag" (report), ``maatregelen" (measures), ``steun" (support) and ``standpunt" (position). The type of language most typical of the conversational data marks its more informal and interactional nature, with interjections like ``oh", ``uh" and ``uhm" (the latter two delay markers), polar response particles like ``mm" (yes) and ``nee" (no), and pronouns like ``ge", ``gij" and ``m'n" (my). These are exactly the kind of interactional tools we saw in §5.1 above: little words that are frequently used and that streamline conversation.

One implication of this is that many of the short and highly frequent turn formats in conversational speech are not well represented in ASR training data, with detrimental consequences for the ability of ASR models to deal with conversational speech. And indeed there are indications that ASR performs less well for such data. One study comparing Google, Microsoft and HuggingFace ASR models for Swedish found that ``for all spontaneous speech, the ASRs frequently fail to produce a transcription for short utterances" \cite{cumbal_you_2021}. In a study comparing gold standard human transcripts with ASR output, it is precisely the words that serve as continuers, feedback signals and other metacommunicative signals that are most frequently missed \cite{zayatsDisfluenciesHumanSpeech2019}.

Missing or incorrectly transcribing short utterances may not be a big problem for speech recognition models whose main function is to deal with relatively clean recordings of non-conversational speech (such as speeches, radio programs, parliamentary meetings and other highly institutionalized text types). But it does spell serious trouble for the use of speech recognition in more interactive contexts such as voice user interfaces and conversational agents. If one goal of ASR is to be able to deal with human speech input in interactive situations, then the current training data may not optimally prepare it for this job.

Recent work in two areas has claimed some territory here. Small corpus size need not be a problem: there are promising ASR results for corpora that amount to only an hour of speech, albeit non-interactive \cite{tyers2021shall}. Further, under the banner of `textless' ASR, Nguyen et al. \shortcite{nguyenGenerativeSpokenDialogue2022} investigate how speech features can be learned from conversational English telephone data using a pipeline that relies less on textual representations. Such techniques may be extended to other types of conversational data, although this inevitably requires dealing with hurdles like noise and non-separate channels. 

Indeed a serious challenge for wider reliance on conversational corpora is that the speech signal in such corpora is in many cases not as pristine as typical ASR data: it includes overlapping speech, non-speech, background noises and comes with all of the complexities generated by different recording environments and equipments. However, to the extent this is a problem, we submit that this is not so much an issue to be solved on the input side, but a matter of ecological validity. If we want ASR systems that are able to deal with the contingencies and exigencies of human interaction in its natural environment \cite{baumann2017recognising,riviereUnsupervisedLearningSpeech2021}, then we better make sure the training data prepare them for this. We anticipate that carefully time-aligned corpora of the type we curate and describe here will play a key role in this. Very likely, selective augmentation of ASR training corpora will provide a scaleable partial solution: including truly interactional data in ASR training sets will meaningfully improve speech recognition in challenging interactional contexts.

\section{Discussion}
\noindent
Conversational corpora are crucial for furthering our understanding of language in its most natural habitat and for building diversity-aware language technology. An increasing number of such corpora is available as a result of decades of concerted efforts in the field of language documentation to compile and preserve linguistic primary data \cite{himmelmannLanguageDocumentationWhat2006,seifartLanguageDocumentationTwentyfive2018}. 

Yet even  though interactional data is gaining prominence, it is still not standard fare in corpus linguistics and NLP research. This results in a kind of double bind: conversational data is hard to find and clear examples of the importance of such data for language science and technology are rare. One of our aims is to free the field from this double bind by compiling open data about available resources, specifying key properties of an interoperable format for interactional data, providing elements of an open data processing pipeline, and pointing out promising research directions. However, this work also comes with ethical considerations, and this is what we discuss first. 

\subsection{Ethical considerations}
\noindent
Ethical considerations start with the question of where to focus our efforts and what kind of research to pursue. The field of natural language processing is famously data-hungry, and current incentives are aligned to harvest more of the same: large amounts of mostly monologic textual data \cite{benderDangersStochasticParrots2021}. In the face of this, it may seem quaint to focus on linguistically diverse, hard-won, small to medium-sized corpora of conversational interaction with relatively few opportunities for the rapid scaling up we have come to expect from web-scrapeable data. But we believe this is a worthwhile, even essential direction for language science and technology. Serious engagement with the cultural, linguistic and interactional diversity embedded in conversational corpora can be a step towards decolonizing the computational sciences \cite{birhaneDecolonisingComputationalSciences2021,birdDecolonisingSpeechLanguage2020}. A true understanding of the workings of language in interaction requires deep engagement with this kind of data \cite{dingemanseTextTalkHarnessing2022}. For anyone seeking to model open-domain conversation, conversational corpora offer a richer and more challenging model than threaded forum posts. For engineers working on conversational agents, actual records of conversation are the best place to study the foundations of interactional infrastructure. For conversation designers, there is no better place to appreciate the sheer flexibility and open-endedness of human interaction than records of people talking. 

Ethical considerations extend to the data sourcing and curation process \cite{rogersChangingWorldChanging2021}. Language resources can have helpful uses but also harmful ones. Language documentation corpora come with their own possibilities and risks for such \textit{dual use} \cite{hovySocialImpactNatural2016,levowDevelopingSharedTask2021}. For datasets archived with community consent and made openly available, there is always the possibility of secondary uses of datasets not foreseen by compilers and communities \cite{seyfeddinipurPublicAccessResearch2019,rogersJustWhatYou2021}. This may include applications such as the possibility of using conversational data to inform pragmatic typology or to improve speech recognition technology, as we have shown above. While neither of these use cases relies on personally identifiable information, more questionable uses might also possible. For instance, audio and video data necessarily includes personally identifiable information (people's voices and likenesses), and the content of quotidian conversations may occassionally feature information that may be subject to privacy considerations. 

For this reason and others, data sourcing and distribution has to be regulated. In the case of language documentation corpora, an important part of this responsibility is carried by language resource archives, which typically provide data use agreements, and by corpus compilers, who typically record the sources, goals and circumstances of data collection as well as point out limitations and restrictions on use \cite{seyfeddinipurPublicAccessResearch2019}. The field of language documentation has long emphasized the importance of vigilance about ethical data collection, informed consent and reproducible research \cite{dwyerEthicsPracticalitiesCooperative2006,bowernLinguisticFieldworkPractical2007,berez-kroekerReproducibleResearchLinguistics2018,goodEthicsLanguageDocumentation2018}. The use of data sheets for NLP data sets \cite{gebruDatasheetsDatasets2021} represents an important convergent development, and may provide inspiration to language documentation archives and corpus compilers. 

Two complicating factors here are worth noting. First, it is the very nature of secondary uses that they cannot be fully foreseen at the point of data collection. This makes it all the more important that researchers are rooted in the communities they work with, and that they clearly communicate the possible implications of having research data archived in internet-accessible repositories \cite{levowDevelopingSharedTask2021}. Second, ethical notions cannot be assumed to be culturally neutral; for instance, Ameka \& Terkourafi point out that Western ethical frameworks privilege autonomy and privacy, whereas ``in some communities, research participants are happy, indeed expect, to be fully identified" \cite{amekaWhatIfImagining2019}. As they note, ideally, questions about data archiving including anonymization practices should be informed by \textit{local} ethical standards. Throughout, a guiding principle should be that data is archived, and its availability and conditions on reuse set, in accordance with the wishes of participants and communities \cite{nathanAccessAccessibilityELAR2013}.

\subsection{Conclusions}
\noindent
We have documented here a first effort at sourcing a maximally diverse set of openly available conversational corpora. We publish an up to date survey of available corpora that provides a headstart for people looking to work with more diverse data sets. And we share specifications and code for an analysis pipeline to enable others to use similar methods for building and curating conversational corpora. In order to further the goal of resource creation, we have also formulated some simple criteria for creating minimally viable conversational corpora.

Having generalisable methods and data representations for dealing with interactional data serves the interest of many communities working with language resources. It contributes towards alleviating problems of resource inequality in two ways: by making visible the relative diversity of corpora already available, and by showing how such data can be productively used. Conversational data enables new research questions in linguistic typology and brings into view new applications for language technology. In time, the increasing availability of interactional data in interoperable formats will provide the foundations for novel work at the intersection of language resources and human language technologies.

\section{Acknowledgements}
\noindent
Funding for the work reported here comes from Dutch Research Council grant NWO 016.vidi.185.205 to MD. We thank Ada Lopez for work on the processing pipeline and Matheus Azevedo for help in collating corpus metadata.

\section{Bibliographical References}\label{reference}

\bibliographystyle{lrec}
\bibliography{ElPaCo_shared_library.bib, additional-refs.bib}

\section{Language Resource References\protect\footnote{Language resources are cited here according to the latest LREC citation style. We note that this style makes it hard to cite web resources in the way recommended by archives. To make available both the archive name (e.g., Endangered Language Archive) as well as a durable URL, we have resorted to a workaround: we encode the URL as an href attribute in the number field. Full and correct .bib metadata is available in our repository.}}

\label{lr:ref}
\bibliographystylelanguageresource{lrec}
\bibliographylanguageresource{language_resources.bib,ElPaCo_shared_library.bib}

\clearpage

\onecolumn
\section*{Appendices}

\subsection{Data access}
\noindent
All included datasets are available to the research community, but most do not allow direct redistribution. Further, conversational data such as this is subject to important ethical considerations (see §6.1). We provide a detailed overview of name, location, as well as authorship and usage rights information for each dataset at \href{https://osf.io/cwvbe}{osf.io/cwvbe}. We plan to maintain this repository of information about open datasets of conversational speech as a resource for anyone interested in compiling a dataset of similar nature or in reproducing the dataset described in this paper. Many of the datasets are distributed by language documentation archives that require the creation of a dedicated user account as well as the signing of a data use agreement. For this reason we are not able to provide direct access to the full dataset directly. 

\subsection{Tools for processing conversational data}
\noindent
Language resource platforms provide files in a range of formats, all of which need parsing to access transcription content. Language documentation corpora often come as ELAN .eaf files, an XML-based format that links utterance transcription content to timing information (with varying levels of precision). Tier structures in ELAN are subject to considerable customization by corpus creators, so they require manual inspection prior to parsing to identify the desired annotation levels. The web-based \href{https://www.gerlingo.com/config_maker.html}{``ELAN inventory" tool} provided by the ARC CoEDL is useful for providing a quick look at the tier structure of ELAN files.

Other common formats of conversational speech corpora are the Talkbank CHAT .cha format, Praat .TextGrid \cite{boersmaPraatDoingPhonetics2013}, and Exmeralda .exb \cite{schmidtEXMARaLDA2014}. This whole range of formats is usually accompanied by parsing tools for various operating systems and scripting languages. We use the Python-based \texttt{scikit-talk} which works with parsers for most formats we encountered while collecting the corpora that make up the current dataset \cite{liesenfeld2021scikit}.

\subsection{List of languages with openly available conversational corpora}
\label{app_languagelist}
\noindent
The table below presents the languages and corpora surveyed in this paper, with glottocodes and families according to Glottolog \cite{hammarstromGlottologGlottologGlottolog2021} and with citations according to source archives. Full details, including corpus statistics, sample annotations and links, are in the study repository at \href{https://osf.io/cwvbe}{osf.io/cwvbe}.  
 
\begin{longtable}{lll}
  \hline
Language (glottocode) & Family & Citation \\ 
  \hline
Akie (mosi1247) & Nilotic & \citelanguageresource{legereCollectionAkie2019} \\ 
  Akpes (akpe1248) & Atlantic-Congo & \citelanguageresource{lauDocumentingAbesabesi2019} \\ 
  Ambel (waig1244) & Austronesian & \citelanguageresource{arnoldDocumentationAmbelAustronesian2017} \\ 
  Anal (anal1239) & Sino-Tibetan & \citelanguageresource{ozerovCommunitydrivenDocumentationNatural2018} \\ 
  Arabic (egyp1253) & Afro-Asiatic & \citelanguageresource{canavanalexandraCALLHOMEEgyptianArabic1997} \\ 
  Arapaho (arap1274) & Algic & \citelanguageresource{cowellConversationalDatabaseArapaho1950} \\ 
  Baa (kwaa1262) & Atlantic-Congo & \citelanguageresource{mollernwadigoDocumentationProjectBaa2016} \\ 
  Besemah (musi1241) & Austronesian & \citelanguageresource{gilDocumentationBesemahMalayic2015} \\ 
  Br.Portuguese (braz1246) & Indo-European & \citelanguageresource{dasilvaProjetoNormaUrbana1996} \\ 
  Bulgarian (bulg1262) & Indo-European & \citelanguageresource{tishevaCorpusSpokenBulgarian} \\ 
  Catalan (stan1289) & Indo-European & \citelanguageresource{garridoGlissandoCorpusMultidisciplinary2013} \\ 
  Chitkuli (chit1279) & Sino-Tibetan & \citelanguageresource{martinezDocumentaryCorpusChhitkulRakchham2020} \\ 
  Cora (sant1424) & Uto-Aztecan & \citelanguageresource{parkerDocumentationCoraSan2020} \\ 
  Croatian (croa1245) & Indo-European & \citelanguageresource{kuvackraljevicCroatianAdultSpoken2016} \\ 
  Czech (czec1258) & Indo-European & \citelanguageresource{ernestusNijmegenCorpusCasual2014a} \\ 
  Danish (dani1285) & Indo-European & \citelanguageresource{wagnerSamtaleBankDanishSpoken2017} \\ 
  Datooga (isim1234) & Nilotic & \citelanguageresource{griscomDocumentationIsimjeegDatooga2018} \\ 
  Dutch (dutc1256) & Indo-European & \citelanguageresource{taalunieCorpusGesprokenNederlands2014} \\ 
  English (nort3314) & Indo-European & \citelanguageresource{canavanalexandraCALLFRIENDAmericanEnglishNonSouthern1996} \\ 
  Farsi (west2369) & Indo-European & \citelanguageresource{canavanalexandraCALLFRIENDFarsiSecond2014} \\ 
  French (stan1290) & Indo-European & \citelanguageresource{torreiraNijmegenCorpusCasual2010} \\ 
  German (stan1295) & Indo-European & \citelanguageresource{canavanalexandraCALLHOMEGermanSpeech1997} \\ 
  Gunwinguu (gunw1252) & Gunwinyguan & \citelanguageresource{aungsiAudioVideoRecordings2014} \\ 
  Gutob (bodo1267) & Austroasiatic & \citelanguageresource{vossDocumentationGrammarGutob2018} \\ 
  Hausa (haus1257) & Afro-Asiatic & \citelanguageresource{caronHausaCollectionCOllections2016} \\ 
  Heyo (heyo1240) & Nuclear Torricelli & \citelanguageresource{diazDocumentationHeyoAuk2018} \\ 
  Hungarian (hung1274) & Uralic & \citelanguageresource{hunyadiHumanhumanHumanmachineCommunication2018} \\ 
  Japanese (nucl1643) & Japonic & \citelanguageresource{nakamuraCABankJapaneseCallFriend2005} \\ 
  Jejueo (jeju1234) & Koreanic & \citelanguageresource{kimMultimodalDocumentationJejuan2018} \\ 
  Juba Creole (suda1237) & Afro-Asiatic & \citelanguageresource{manfrediJubaCreoleCollection2016} \\ 
  Kakabe (kaka1265) & Mande & \citelanguageresource{vydrinaDescriptionDocumentationKakabe2013} \\ 
  Kelabit (kela1258) & Austronesian & \citelanguageresource{hemmingsDocumentationKelabitLanguage2017} \\ 
  Kerinci (keri1250) & Austronesian & \citelanguageresource{fadlulKerinciSungaiPenuh2016} \\ 
  Khinalug (khin1240) & Nakh-Daghestanian & \citelanguageresource{rind-pawlowskiKhinalugDocumentationProject2016} \\ 
  Kichwa (tena1240) & Quechuan & \citelanguageresource{grzechUpperNapoKichwa2020} \\ 
  Korean (kore1280) & Koreanic & \citelanguageresource{canavanalexandraCALLFRIENDKorean1996} \\ 
  Kula (kula1280) & Timor-Alor-Pantar & \citelanguageresource{williamsDocumentingLanguageInteraction2017} \\ 
  Laal (laal1242) & Laal & \citelanguageresource{lionnetLaalLanguageDocumentation2020} \\ 
  Limassa (lima1246) & Atlantic-Congo & \citelanguageresource{winkhartDocumentationRemnantBakaGundi2016} \\ 
  Mambila (came1252) & Atlantic-Congo & \citelanguageresource{ogunsolaDocumentationLenMambila2018} \\ 
  Mandarin (mand1415) & Sino-Tibetan & \citelanguageresource{canavanalexandraCALLHOMEMandarinChinese1996} \\ 
  Minderico (mind1263) & Indo-European & \citelanguageresource{carvalhoferreiraMindericoEndangeredLanguage2011} \\ 
  N|uu (nuuu1241) & Tuu & \citelanguageresource{guldemannTextDocumentationUu2014} \\ 
  Nahuatl (cent2132) & Uto-Aztecan & \citelanguageresource{amithAudioCorpusSierra2009} \\ 
  Nasal (nasa1239) & Austronesian & \citelanguageresource{mcdonnellDocumentationNasalOverlooked2017} \\ 
  Nganasan (ngan1291) & Uralic & \citelanguageresource{brykinaNganasanSpokenLanguage2018} \\ 
  Otomi (esta1236) & Otomanguean & \citelanguageresource{hernandez-greenDocumentationSanJeronimo2009} \\ 
  Pagu (pagu1249) & North Halmahera & \citelanguageresource{hisyamProjectLIPIIndonesian2013} \\ 
  Polish (poli1260) & Indo-European & \citelanguageresource{pezikPELCRAPolishSpoken2011} \\ 
  S.Qiang (sout2728) & Sino-Tibetan & \citelanguageresource{simsDocumentationYongheQiang2018} \\ 
  Saami (pite1240) & Uralic & \citelanguageresource{wilburPiteSaamiDocumenting2009} \\ 
  Sakun (suku1272) & Afro-Asiatic & \citelanguageresource{thomasSakunSukurLanguage2014} \\ 
  Sambas (kend1254) & Austronesian & \citelanguageresource{tadmorLanguagesWesternBorneo2007} \\ 
  Siona (sion1247) & Tucanoan & \citelanguageresource{martineDocumentationEcuadorianSiona2012} \\ 
  Siputhi (swat1243) & Atlantic-Congo & \citelanguageresource{shahMultimediaCorpusSiPhuthi2019} \\ 
  Spanish (stan1288) & Indo-European & \citelanguageresource{canavanalexandraCALLHOMESpanishSpeech1996} \\ 
  Tehuelche (tehu1242) & Chonan & \citelanguageresource{domingoTehuelcheLanguageCollection2019} \\ 
  Totoli (toto1304) & Austronesian & \citelanguageresource{letoCollectionTotoli2010} \\ 
  Tseltal (tzel1254) & Mayan & \citelanguageresource{polianTseltalDocumentationProject2010} \\ 
  Ulwa (ulwa1239) & Misumalpan & \citelanguageresource{barlowDocumentationUlwaEndangered2017} \\ 
  Vamale (vama1243) & Austronesian & \citelanguageresource{rohlederDocumentationDescriptionVamale2018} \\ 
  Wooi (woii1237) & Austronesian & \citelanguageresource{unterladstetterCollectionWooi2013} \\ 
  Yakkha (yakk1236) & Sino-Tibetan & \citelanguageresource{schackowDocumentationGrammaticalDescription2014} \\ 
  Yali (pass1247) & Nuclear Trans New Guinea & \citelanguageresource{riesbergYaliSummitsCollection2015} \\ 
  Yélî Dnye (yele1255) & Yele & \citelanguageresource{levinsonCollectionYeliDnye2019} \\ 
  Zaar (saya1246) & Afro-Asiatic & \citelanguageresource{caronCollectionZaarLangage2014} \\ 
  Zauzou (zauz1238) & Sino-Tibetan & \citelanguageresource{liDocumentationZauzouEndangered2017} \\ 
   \hline
\hline
\end{longtable}

\textit{A note on sign languages.} The table above includes only spoken languages. While we have sampled as broadly as possible, sign language corpora of conversation are still quite rare \citelanguageresource{kopfOverviewDatasetsSign2021,yinIncludingSignedLanguages2021}, and the handful that are openly available are primarily organized in terms of sign-level rather than turn-level annotations. This holds for Chatino Sign Language \citelanguageresource{houDocumentingChatinoSign2018}, Côte d'Ivoire Sign Language \citelanguageresource{tanoDocumentationDescriptionSign2013}, French Belgian Sign Language \citelanguageresource{meurantCorpusLSFBFirst2015}, German Sign Language \citelanguageresource{hankeMEINEDGSOffentliches2020}, and Macau Sign Language \citelanguageresource{yimPreliminaryDocumentationMacau2014}. Although there is ample evidence that sign language conversations are also turn-organized \cite{devosPredictingConversationalTurns2021}, the sign-level annotations mean that additional processing steps would be required to render such corpora interoperable with the minimal data format specifications proposed here. We think this is best done in consultation with corpus compilers and language experts.

\end{document}